# Ranking by Dependence—A Fair Criteria

**Harald Steck**
Computer-Aided Diagnosis and Therapy
Siemens Medical Solutions, 51 Valley Stream Parkway E51
Malvern, PA 19355, USA
*harald.steck@siemens.com*

## Abstract

Estimating the dependences between random variables, and ranking them accordingly, is a prevalent problem in machine learning. Pursuing frequentist and information-theoretic approaches, we first show that the p-value and the mutual information can fail even in simplistic situations. We then propose two conditions for regularizing an estimator of dependence, which leads to a simple yet effective new measure. We discuss its advantages and compare it to well-established model-selection criteria. Apart from that, we derive a simple constraint for regularizing parameter estimates in a graphical model. This results in an analytical approximation for the optimal value of the equivalent sample size, which agrees very well with the more involved Bayesian approach in our experiments.

## 1 Introduction

Quantifying the degree of dependence between two random variables is pivotal to many machine-learning approaches, including discriminative and generative learning. When given the *empirical* distribution as implied by the given data, a properly regularized dependence-measure is essential. In the context of model-selection, various scoring functions have proven useful, including the p-value, Akaike Information Criteria (AIC) [1], Bayesian Information Criteria (BIC) [16], Minimum Description Length (MDL) [15] and (Bayesian) posterior probability. Typically, the objective is to estimate the predictive accuracy of the learned models. For instance, $-$AIC approximates the test error $E_p[-\log \hat{p}]$ in a frequentist way, i.e., the model parameters are based on the *empirical* distribution $\hat{p}$, while the expectation $E$ is with respect to the (unknown) *true* distribution $p$ [1].

As to obtain a fair ranking according to the dependences between pairs of random variables, our objective is to assess dependence with respect to the (unknown) *true* distribution $p$, rather than using the empirical probabilities as model parameters. Like most of the model-selection criteria above, our approach is also based on the log maximum likelihood. As our objective is different, however, we need a new kind of regularization: the two limiting conditions are outlined in Section 3. We present the resulting new measure, which we call *standardized information*, in Section 4. Section 5 discusses its interesting properties, including its particularly simple form, and its relationship to common model-selection criteria. Our experiments in Section 7 show that commonly-used measures can already fail in simplistic situations, while our new measure yields the expected results.

Apart from that, our frequentist approach also yields the optimal regularization of the parameter estimates of a graphical model: we derive an analytical approximation for the optimal value of the equivalent sample size in Section 6. It agrees very well with the (involved) Bayesian approach in our experiments (Section 7.2).

## 2 Notation

This section introduces relevant notation and reviews bias-correction of mutual information. For simplicity and without loss of generality, we focus on *two* random variables following a multinomial distribution, say $A$ and $B$; for the generalization to many variables, see Section 5.5; concerning continuous variables, and the estimation of entropy thereof, paper [3] discusses the relevant issues, which differ from the objective of this paper. Let $a$ denote a state of $A$, and $|A|$ its number of states; $p(A, B)$ is the *true* (yet unknown) distribution, and $\hat{p}(A, B)$ is the *empirical* one, as implied by data $D$: $\hat{p}(A = a, B = b) = N_{ab}/N$, where $N_{ab}$ serve as sufficient statistics (frequencies of $(A = a, B = b)$), and $N$ is the sample size.

Mutual information $I(A,B)$ is the information-theoretic measure of dependence (cf. [5] for an overview).[1] As the true distribution $p(A,B)$ is typically not available in machine-learning applications, the true $I(A,B)$ unfortunately cannot be computed. Instead, it has to be estimated from the *empirical* distribution $\hat{p}$. The frequentist plug-in estimator reads

$$\hat{I}(A,B) = \sum_{a,b} \hat{p}(a,b) \log \frac{\hat{p}(a,b)}{\hat{p}(a)\hat{p}(b)}. \qquad (1)$$

It is well known that this plug-in estimator is biased (e.g., [11, 18, 12]). The bias-corrected estimator reads

$$\hat{I}^{\mathrm{BC}}(A,B) = \hat{I}(A,B) - d_{AB}/(2N) \qquad (2)$$

in leading order in $N$, where $d_{AB}$ are the *degrees of freedom*; if all joint configurations $(A=a, B=b)$ occur in data $D$, $d_{AB} = (|A|-1)(|B|-1)$. If some joint configurations are missing in the (small) data set, the *effective* degrees of freedom are used instead, cf. [4].

## 3 Two Conditions for Regularization

In this section we propose two basic conditions that a well-regularized—and thus fair—estimator of dependence has to fulfill. These conditions apply to the extreme cases of vanishing and maximum dependence.

### 3.1 First Limiting Condition

Before we present the first condition, namely for the case of a vanishing degree of dependence, let us first consider a simple example for illustration purposes.

Consider the task of ranking two pairs of variables, $(A,B)$ and $(X,Y)$, according to their degree of dependence (cf. Figure 1). Assume that the given data implies $\hat{I}(A,B) < \hat{I}(X,Y)$. This suggests from a naive point of view that $X$ and $Y$ are more strongly dependent than $A$ and $B$.

As the various discrete variables may have different numbers of states, consider the case where $d_{AB} < d_{XY}$. As implied by Fig. 1, assume the bias-corrected estimator (cf. Eq. 2) yields $\hat{I}^{\mathrm{BC}}(A,B) < \hat{I}^{\mathrm{BC}}(X,Y)$, which also suggests that the dependence between $X$ and $Y$ is stronger than the one between $A$ and $B$.

Adopting a frequentist point of view, the estimator for mutual information follows a distribution. While the bias-correction accounted for its mean, let us now also account for its standard deviation: as implied by Fig. 1, assume that $\hat{I}^{\mathrm{BC}}(A,B) > \mathrm{std}(\hat{I}(A,B))$, while $\hat{I}^{\mathrm{BC}}(X,Y) < \mathrm{std}(\hat{I}(X,Y))$. Obviously, when the deviation of $\hat{I}$ from mean $d/(2N)$ is measured *in terms of the standard deviation*, then $\hat{I}(X,Y)$ is closer to the mean $d_{XY}/(2N)$ than is $\hat{I}(A,B)$ to $d_{AB}/(2N)$. This implies that $(A,B)$ are more strongly dependent than are $(X,Y)$, which is contrary to the previous results.

[1] As we focus on two variables, we use log (maximum) likelihood ratio and mutual information interchangeably.

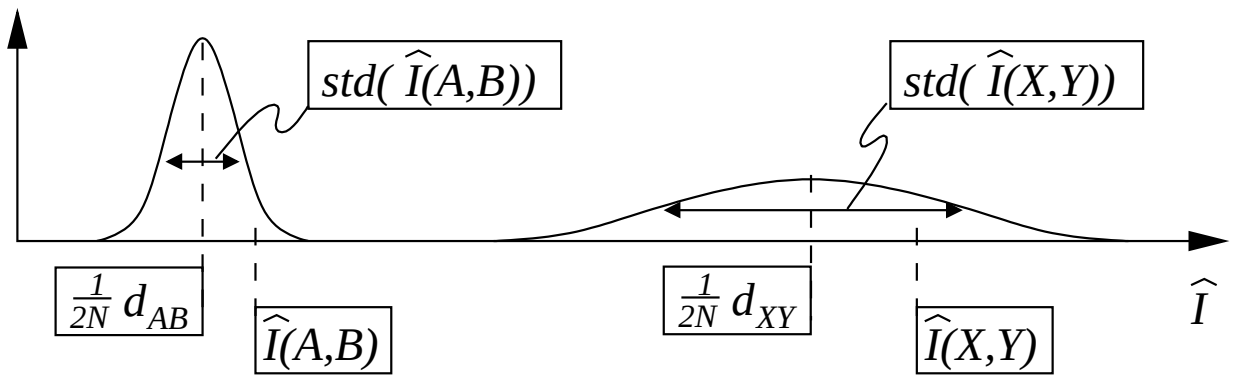

Figure 1: Example: expected distribution of the plug-in estimator $\hat{I}$ for variables $(A,B)$ versus $(X,Y)$.

In summary, this example illustrates that a fair assessment of dependence requires that the *same scale* is used for both pairs $(A,B)$ and $(X,Y)$. The standard deviation is such a natural length scale.

As to formalize this insight, let us focus on the limiting case of a vanishing dependence in this section (cf. next section for strong dependence): $I \to 0$, i.e., the variables are independent w.r.t. the (unknown) *true* distribution $p$. In this case, like in basic frequentist independence testing, the standard deviation of the estimator $\hat{I}$ is given by

$$\mathrm{std}\left(\hat{I}(A,B)\right) = \frac{\sqrt{d_{AB}}}{\sqrt{2}N} + \mathcal{O}(\frac{1}{N^{3/2}}). \qquad (3)$$

Note that we do not need to calculate the standard deviation for the general case of possibly strong dependence here (cf. [12, 11] for elaborations on that), rendering our approach much simpler and computationally much more efficient. Now we can formulate
**Condition 1:**
*In the limiting case of vanishing dependence, a well-regularized estimator for dependence, $\hat{J}$, has to fulfill*

$$\hat{J}(A,B) \approx \hat{R}(A,B) \equiv \frac{\hat{I}(A,B) - \frac{d_{AB}}{2N}}{\sqrt{d_{AB}/2}/N} \qquad (4)$$

*when $\hat{I}(A,B) \approx d_{AB}/(2N)$; more formally $\hat{J} \to \hat{R}$ and $d\hat{J}/d\hat{I} \to d\hat{R}/d\hat{I}$ as $\hat{I}(A,B) \to d_{AB}/(2N)$ for fixed $d_{AB} > 0$, ignoring higher order derivatives.*

This condition can be interpreted in various ways: first, Eq. 4 mimics the behavior of the p-value, which also accounts for the variance of the distribution under the independence-assumption, cf. Section 5.3; second, the ratio in Eq. 4 can be viewed as the one-dimensional version of the well-known Mahalanobis distance, which is typically used in the context of Gaussian distributed variables; third, the ratio in Eq. 4 may be interpreted as the well-known standardized score or z-score, which is usually used to assess the deviation of individual data points with respect to a given distribution.

### 3.2 Second Limiting Condition

As to ensure consistency with information theory, we also require **Condition 2:** *If the random variables $A$ and $B$ are dependent based on the (unknown)* true *distribution $p$ (i.e., $I(A,B) > 0$), then a well-regularized estimator for dependence, $\hat{J}$, has to fulfill*

$$\hat{J}(A,B) \to f(\hat{I}(A,B)) \quad \text{as} \quad N \to \infty, \tag{5}$$

*where $f$ is a strictly monotonically growing function that is independent of $d_{AB}$.*

This condition essentially applies to the cases where the estimated mutual information is large, i.e., where $\hat{I}(A,B) \gg d_{AB}/(2N)$. Obviously, the ratio in Eq. 4 of condition 1 does not fulfill condition 2 due to $d_{AB}$ in the denominator. Note that we require for $f$ only independence from $d_{AB}$, but not from $N$.

## 4 New Regularized Estimator

This section outlines the estimators that comply with the two conditions proposed above. While such an estimator $\hat{J}$ can take a multitude of functional forms, the most important distinction is as to how 'quickly' they transition from following the behavior required for small values $\hat{I}$ (condition 1) to the behavior for large values $\hat{I}$ (condition 2). It is easy to construct a family of functions where the rate of this transition is controlled by a free parameter.

In the following we focus on this transition being of the same order as the standard deviation of the distribution of $\hat{I}$ under the independence-assumption, which is a natural length-scale for the transition (cf. also Section 5.4). The following estimator takes a particularly elegant form:

$$\widehat{SI}(A,B) = \sqrt{2N\hat{I}(A,B)} - \sqrt{d_{AB}}, \tag{6}$$

as it differs from the commonly-used bias-corrected estimator for mutual information only in the square-roots of each *individual* term. The proof that $\widehat{SI}$ indeed obeys both the conditions proposed in Section 3 is obvious: condition 1 holds because

$$\hat{R}(A,B) = \frac{\hat{I}(A,B) - d_{AB}/(2N)}{\sqrt{d_{AB}/2}/N} =$$

$$\left\{\sqrt{2N\hat{I}(A,B)} - \sqrt{d_{AB}}\right\} \frac{\left\{\sqrt{2N\hat{I}(A,B)} + \sqrt{d_{AB}}\right\}}{\sqrt{2d_{AB}}}$$

$$= \widehat{SI}(A,B) \cdot \sqrt{2} \cdot \left\{1 + \frac{\widehat{SI}(A,B)}{2\sqrt{d_{AB}}}\right\}$$

shows that, as $\hat{I}(A,B) \to d_{AB}/(2N)$, $\hat{R}(A,B)$ and $\widehat{SI}(A,B)$ are identical up to linear order (ignoring the irrelevant constant $\sqrt{2}$). Also Condition 2 holds: when $N \to \infty$ and $I(A,B) > 0$, then $\widehat{SI}(A,B)/\sqrt{2N} \to \sqrt{\hat{I}(A,B)} \to \sqrt{I(A,B)} > 0$, which is strictly monotonically growing in $\hat{I}(A,B)$ and independent of $d_{AB}$. This proof also shows that the length-scale of the transition of $\widehat{SI}$ between conditions 1 and 2 is of the order of the standard deviation, as desired. We call $\widehat{SI}$ the *standardized information*, as it measures weak dependences in (fractional) multiples of the *standard deviation*, which accounts for its regularization.

## 5 Discussion of Estimator

This section discusses interesting properties of the presented *standardized information* (cf. Eq. 6), as well as the differences to popular model-selection criteria.

### 5.1 Ranking by Dependence

The two conditions proposed in Section 3 ensure that dependence is assessed on a scale that is invariant under different cardinalities of the variables involved. This leads to a fair comparison—and thus ranking—of variables with different numbers of states. Without condition 1, weak dependences would be overestimated for variables with *large* cardinalities; with condition 1 but without 2, strong dependences would be overestimated for variables with *small* cardinalities.

This invariance of the scale is the main difference to other popular model-selection criteria. For instance, AIC, BIC or MDL contain only an *additive* correction (the penalty term for complexity) to the log likelihood ratio term. This has two consequences. First, as their penalty terms are typically larger than the bias-correction in Eq. 2, the zero-point of the resulting 'scale of dependence' gets shifted by more, resulting in an increased penalty for variables with large cardinalities compared to our measure. Second, these scores lack a re-scaling factor that would render the scale of dependence invariant under varying cardinalities of the variables. Therefore, these scores are appropriate for determining whether a dependence (or an edge in a graphical model) is significant/notable due to thresholding (additive correction), but they do not necessarily yield a fair quantitative ranking.

When the (unknown) true $I$ is small, then $\widehat{SI}$ can also be expected to take a small positive value (cf. bias correction). Only if $\widehat{SI} > c$ for some threshold value $c > 0$, then the dependence appears to be notable/significant. Using Fisher's asymptotic result [7], $c$ is related to the p-value $\alpha$ according to $\Phi(\sqrt{2}c) \approx 1-\alpha$, where $\Phi$ is the cumulative function of the standard normal distribution (cf. also Sect. 5.4).

### 5.2 Estimating the Conditional Entropy

In contrast to generative learning (of a joint distribution over variables), discriminative approaches often aim to learn the optimal *conditional* probability distribution $p(Y|X)$, where $Y$ is the class variable and $X$ is a (vector of) input variables/features. While minimizing the unknown true conditional entropy $H(Y|X)$ is a valid goal, the maximum likelihood (ML) estimate of the conditional entropy $\hat{H}(Y|X)$ is typically minimized in practice. While this criteria seems different from maximizing the ML estimate of the mutual information $\hat{I}(Y,X)$ at first glance, both criteria are in fact equivalent. This is obvious from the fact that

$$\hat{I}(Y,X) = \hat{H}(Y) - \hat{H}(Y|X), \tag{7}$$

where $\hat{H}(Y)$ is a constant, as it is the entropy of the class variable $Y$ based on the empirical distribution $\hat{p}(Y)$, which is solely determined by the given data, and it is thus independent of the input variables $X$.

From a slightly different view point, one can interpret both $\hat{H}(Y)$ and $\hat{H}(Y|X)$ to be estimated conditional entropies, where $Y$ and $X$ are assumed independent in the former case, while dependent in the latter. Hence, the mutual information $\hat{I}(Y,X)$ can be interpreted as the difference in the discriminative scores (conditional entropies) of the two models where $Y$ and $X$ are independent vs. dependent (cf. [5] for other/standard interpretations of $I$).

This is a particular example of the general concept of *absolute* and *relative* scores: absolute scores (e.g., $\hat{H}(Y|X)$ here) refer to values assigned to individual models, while a relative score (e.g., $\hat{I}(Y,X)$ here) is the difference in the (absolute) scores between two models. Even though it is irrelevant for the optimization task whether the absolute or relative score is used, the estimator of the *relative* score often has a major advantage (at least in the frequentist approach): it can more easily be regularized because its distribution (e.g., mean and variance, ignoring higher-order moments) can be determined more easily. For this reason, we prefer to use the well-regularized (relative) score $\widehat{SI}$ in Eq. 6 instead of the unregularized (absolute) score $\hat{H}(Y|X)$.

### 5.3 Numerical Instability of P-Value

The p-value originates from frequentist hypothesis testing. The p-value of $\hat{I}(A,B)$ is the probability $p(K \geq \hat{I}(A,B))$, where the random variable $K$ follows the distribution of the mutual-information statistic under the assumption that $A$ and $B$ are independent. It is well-known that, in the asymptotic limit, $2NK$ follows a $\chi^2$-distribution with $d_{AB}$ degrees of freedom, which has mean $d_{AB}$ and variance $2d_{AB}$, consistent with the results above. Note that the p-value—as it is based on a distribution—in particular accounts for both the mean and the variance (not to mention higher-order moments), like $\widehat{SI}$. Despite this desirable regularization-property of the p-value, the numerical computation of the p-value often becomes instable. The reason is that $2N\hat{I}$ takes values in the infinite range $[0,\infty)$, which then gets mapped to the unit interval [0,1] for the p-value: large values of $2N\hat{I}$ map to values close to 1 of the cumulative density function, and thus suffer from rounding errors due to limited numerical accuracy, which is also apparent in our experiments in Section 7.

### 5.4 Gaussian Approximation

As mentioned by Fisher [7], a random variable $L \sim \chi^2_d$ following a $\chi^2$-distribution with $d$ degrees of freedom can be transformed into one following a normal distribution approximately: $\sqrt{L} - \sqrt{d} \ \sim \mathcal{N}(0,1/2)$ in the limit $d \to \infty$. Note that the approximately normal distribution of the transformed variable is independent of the degrees of freedom $d$, thus providing a standardized scale for assessing dependence. This shows that Fisher's transformation can be understood as a derivation alternative to ours in Section 4. Apart from that, this agreement also confirms that in our approach in Section 4, the transition between the two proposed conditions indeed takes place at the correct length-scale. Moreover, Fisher noted that replacing $\sqrt{d}$ by $\sqrt{d - 1/2}$ yields an improved approximation accuracy for finite values of $d$ [7].

### 5.5 Generalization to Many Variables

While we have focused on the joint distribution of only two variables so far, now we briefly sketch the various ways of generalizing our standardized information.

First, the standardized information $\widehat{SI}$ (Eq. 6) can be easily generalized to a (conditional) distribution over a *vector of random variables*: this generalization is analogous to the one of the log likelihood ratio, or the one of mutual/multi information, e.g., see [5, 4].

Second, when learning graphical models containing many random variables/nodes (see e.g. [6] for an overview), many scoring functions comprise the log maximum likelihood term and a penalty for model complexity; they are optimized by means of a (heuristic) search strategy. Given complete data, these scores decompose into a sum of 'local' scores pertaining to a small number of variables only (according to the graph structure). Each of these 'local' scores reflects a conditional (in-)dependence. Our results also carry over to each of these local scores, and hence allow one to rank the various (conditional) dependencies between the different variables in the graph in a fair way.

### 5.6 Normalized Mutual Information

In practical applications, like in medical image processing or bioinformatics (e.g., [14, 21]), the *normalized mutual information* is a widely used score, even though it is theoretically not well understood. There are two insights due to our approach, which help shed light onto why the normalized mutual information works fine in some applications, while it fails in others. It takes values in $[0, 1]$ and reads

$$\widehat{NI}(A,B) = \frac{\hat{I}(A,B)}{\left(\hat{H}(A) + \hat{H}(B)\right)/2}. \quad (8)$$

Similar to condition 1 in our approach, $\widehat{NI}$ in the denominator of Eq. 8 uses the 'width' of the distribution under the independence-assumption, under which the joint entropy decomposes like $\hat{H}(A,B) = \hat{H}(A) + \hat{H}(B)$. Thus, $\widehat{NI}(A,B)$ obviously measures $\hat{I}(A,B)$ in (fractional) multiples of the empirical marginal entropies $\hat{H}$; $\hat{H}$ may be viewed as the information-theoretic measure for the 'width' of a distribution, in place of standard deviation in frequentist statistics.

Besides these analogies, there are also three major difference to our approach. First, $\hat{H}(A) + \hat{H}(B)$ yields the widths of the marginal distributions of $A$ and $B$ instead of the distribution of the mutual-information statistic itself. Note, however, that $\hat{H}$ typically also increases with the number of states of $A$ and $B$, although at a different rate than $\sqrt{d_{AB}}$ in the denominator in Eq. 4. Second, unlike in Eq. 4, there is no bias-correction term in $\widehat{NI}$. Compared to the (unregularized) plug-in estimator for mutual information, however, the partial regularization of $\widehat{NI}$ due to the denominator may yield more robust results for small data sets than the plug-in estimator $\hat{I}$. This is indeed observed in practical applications (e.g., [14, 21]), which may explain its popularity among practitioners. Third, in $\widehat{NI}$ the regularization does *not* decrease as the sample size increases, i.e., it does not obey the second condition given in Section 3.2. As a consequence, the dependence between variables with a large number of states tends to be underestimated when given large data sets, and $\widehat{NI}$ is not a suitable measure of dependence in this regime.

## 6 Regularization of Parameters

In the context of (graphical) model selection, the Bayesian approach allows one to regularize the model structure (sparseness of the graph) as well as the model parameters (smoothing of the probabilities), based on the researcher's choice of the prior distribution. In contrast, frequentist approaches typically focus on restricting model complexity while using the (unregularized) ML estimates of the parameters. In the following, we outline a simple frequentist approach for determining the optimal regularization of parameter estimates.

First, we have to assume a reasonable functional form of the regularized parameter estimates, say $\tilde{p}$. A common choice is[2]

$$\tilde{p}_{ab} = \frac{N_{ab} + N' \cdot q_{ab}}{N + N'},$$

where $N' \cdot q_{ab}$ represents additional virtual counts in the various joint states: $N'$ is the overall number and $q_{ab}$ denotes a probability distribution; in the absence of background knowledge, the (prior) distribution $q$ is typically chosen to be uniform. While this is a plausible assumption from a frequentist perspective, it is also obtained in the Bayesian approach when assuming a Dirichlet prior over the model parameters and averaging them out [10]. In the Bayesian approach, $N'$ is the scale parameter of the Dirichlet prior, and is also called the equivalent sample size. As it has more or less counter-intuitive effects on the structure-learning results [19], choosing a 'good' value for $N'$ is crucial. Cross-validation or maximized (Bayesian) marginal likelihood are two standard ways for determining the value of $N'$, e.g., [19].

In the following, we propose an alternative approach, namely a simple frequentist constraint determining the value of $N'$. As to simplify subsequent calculations, we consider a quantity that is linear in the (unknown) true distribution $p$, namely $\sum_{a,b} p(a,b) \log[\hat{p}(a,b)/(\hat{p}(a)\hat{p}(b))]$.[3] We estimate its value by using the empirical mutual information $\hat{I}$. As before, one can easily obtain the approximate leading-order bias-correction [1], and obtain the constraint

$$\sum_{a,b} \tilde{p}_{ab} \log \frac{\hat{p}(a,b)}{\hat{p}(a)\hat{p}(b)} = \hat{I}(A,B) - \frac{d_{AB}}{N} + \mathcal{O}(\frac{1}{N^2}), \quad (9)$$

where we have plugged in the regularized parameter estimates $\tilde{p}_{ab}$ in place of the (unknown) true probabilities $p(a,b)$. As the terms on the right hand side are determined by the given data, obviously the regularized estimates $\tilde{p}_{ab}$—and thus $N'$—are determined by this equality-constraint. Note that the left hand side in Eq. 9 is monotonically decreasing with growing $N'$ if the distribution $q$ is uniform.

Moreover, as Eq. 9 is linear in the regularized probabilities $\tilde{p}_{ab}$, this immediately yields an approximate solution for $N'$: assuming that $N'^2 \ll N$, which is typically the case in real-world applications, we take the first-order Taylor expansion of $\tilde{p}_{ab}$ with respect to $N'$ at $N' = 0$, namely $\tilde{p}_{ab} = \hat{p}(a, b) + N'(q_{ab} - N_{ab}/N)/N + \mathcal{O}(N'^2/N^2)$, and insert it into Eq. 9, which can then easily be solved for $N'$:

$$N' = \frac{d_{AB}}{\hat{I}(A, B) - \langle \log \frac{\hat{p}(a,b)}{\hat{p}(a)\hat{p}(b)} \rangle_q} + \mathcal{O}(\frac{N'^2}{N}) \qquad (10)$$

where $\langle \cdot \rangle_q$ denotes the expectation with respect to the (prior) distribution $q$; it yields a non-positive number if the (prior) distribution $q$ is chosen to be uniform (a direct consequence of the ML principle); hence, $N'$ is indeed a positive number, as expected.

[2] As before, we consider only a pair of variables here for simplicity but without loss of generality.

[3] We assume $\hat{p}(a,b) > 0$ for all $a, b$; otherwise we replace $\hat{p}(a,b)$ by $\hat{p}^+(a,b) \equiv \max\{N_{ab}, 1\}/N$, as to avoid numerical instabilities.

| B \ A | 1 | 2 | 3 | 4 |
|---|---|---|---|---|
| 1 | .1+z | .1−z | .025 | .025 |
| 2 | .1−z | .1+z | .025 | .025 |
| 3 | .025 | .025 | .1+z | .1−z |
| 4 | .025 | .025 | .1−z | .1+z |

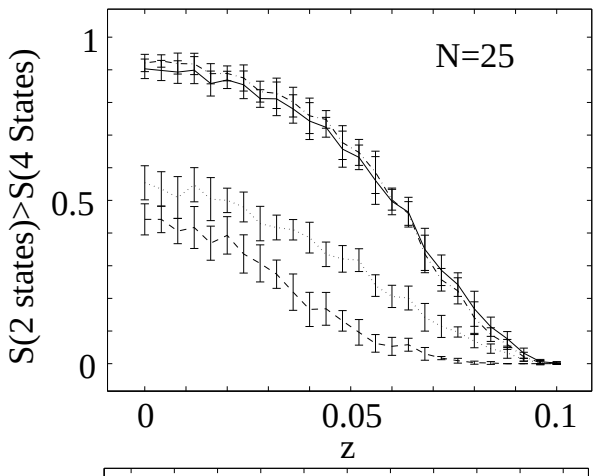

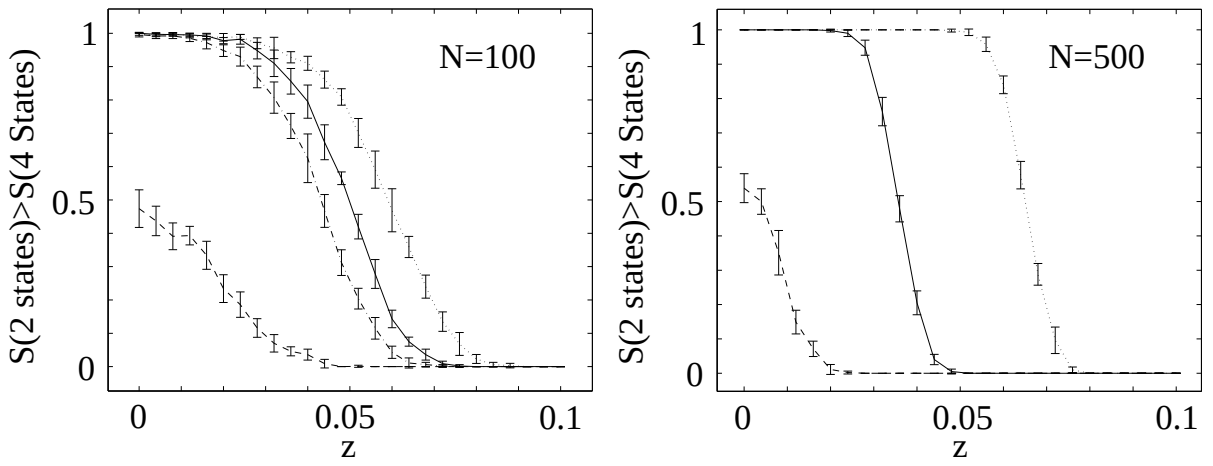

Figure 2: Discretization: from the true distribution (cf. table), samples of sizes $N = 25, 100$ and 500 are sampled. Based on 100 re-samples for each $N$, the fraction where the dependence-measure favours 2 states over 4 states is shown along the y-axis of the graphs; the dependence between $A$ and $B$ increases along the x-axis, parameterized by $z \in [0, .1]$. The four measures are $\hat{I}^{\text{BC}}$ (dashed line), $\widehat{SI}$ (solid), $\widehat{NI}$ (dotted) and p-value (dash-dotted).

A few comments on Eq. 10 are in order. The first interesting insight is that $N'$ is independent of the sample size $N$, i.e., once $N'$ has been determined for a given data set, it does not need to be adapted when additional data (from the same distribution) becomes available. This also suggests that our assumption $N'^2 \ll N$ can indeed be expected to hold for a sufficiently large sample size $N$. Second, the generalization to more than two variables is analogous to Section 5.5. Third, in our experiments the results of this simple constraint are consistent with the ones obtained using the two established, but computationally more expensive methods (Section 7.2).

# 7 Experiments

The first section illustrates the advantages of our new standardized-information $\widehat{SI}$ in two application areas: discretization and feature selection. The second section illustrates regularization of parameter-estimates.

## 7.1 Standardized Information Measure

In this section, we show that established measures, like the bias-corrected mutual information $\hat{I}^{\text{BC}}$, the p-value (based on the $\chi^2$-distribution) and the normalized mutual information $\widehat{NI}$, fail already in simplistic experiments, whereas our measure $\widehat{SI}$ (cf. Eq. 6) proves robust. We use data sampled from a known distribution so that the ground-truth is available for evaluation of the various dependence-measures.

In our first experiment, we consider two discrete random variables with four states each, following the distribution shown in the table in Fig. 2. The free parameter $z \in [0, .1]$ can be varied as to change the degree of dependence between the two variables $A$ and $B$. The objective is to 'discretize' the variables $A$ and $B$, i.e., to determine whether the effective number of states is 2 or 4.[4] The desired behavior obviously is a follows: when $z = 0$, then the (true) $I(A, B) = 0$, and hence the estimation from *finite data* should favour 2 states, cf. also Occam's razor. However, when $z = .1$, then 4 states should be preferred. The transition between these two extremes should take place at smaller values $z$ as the sample size $N$ increases, cf. again Occam's razor. Figure 2 summarizes our results: the p-value as well as our new standardized information $\widehat{SI}$ show exactly the desired behavior. The only difference is that the transition tends to occur at slightly higher $z-$values for $\widehat{SI}$ than for the p-value. At sample sizes $N \geq 200$, the p-value approaches zero and fails due to the limited numerical accuracy available (in MATLAB); no results concerning the p-value were thus obtained for $N = 500$ in Figure 2.

Figure 2 also shows that the bias-corrected mutual-information does not provide sufficient regularization: when $z = 0$, $\hat{I}^{BC}$ is essentially undecided between the two hypotheses (2 states vs. 4 states), although the simpler hypothesis should be favoured in this case (cf. Occam's razor). $\hat{I}^{BC}$ hence tends to favour too large a number of states.

The results regarding $\widehat{NI}$ lie between the ones for $\hat{I}^{\text{BC}}$ and the p-value given a small sample size $N = 25$. This suggests that regularization by means of the standard deviation may be more effective than the one relying on the bias in this regime. As the sample size $N$ in-creases (cf. $N = 100$, $500$ in Figure 2), however, the degree of regularization does not diminish in $\widehat{NI}$, as discussed in Section 5. This explains why the transition from favouring 2 states over 4 states stays put at rather large $z$-values when the sample size increases (cf. $N = 500$ in Fig. 2), which is not desirable.

[4] $(A, B)$ are chosen to be already discrete for clarity here; see [20] for discretization of continuous variables.

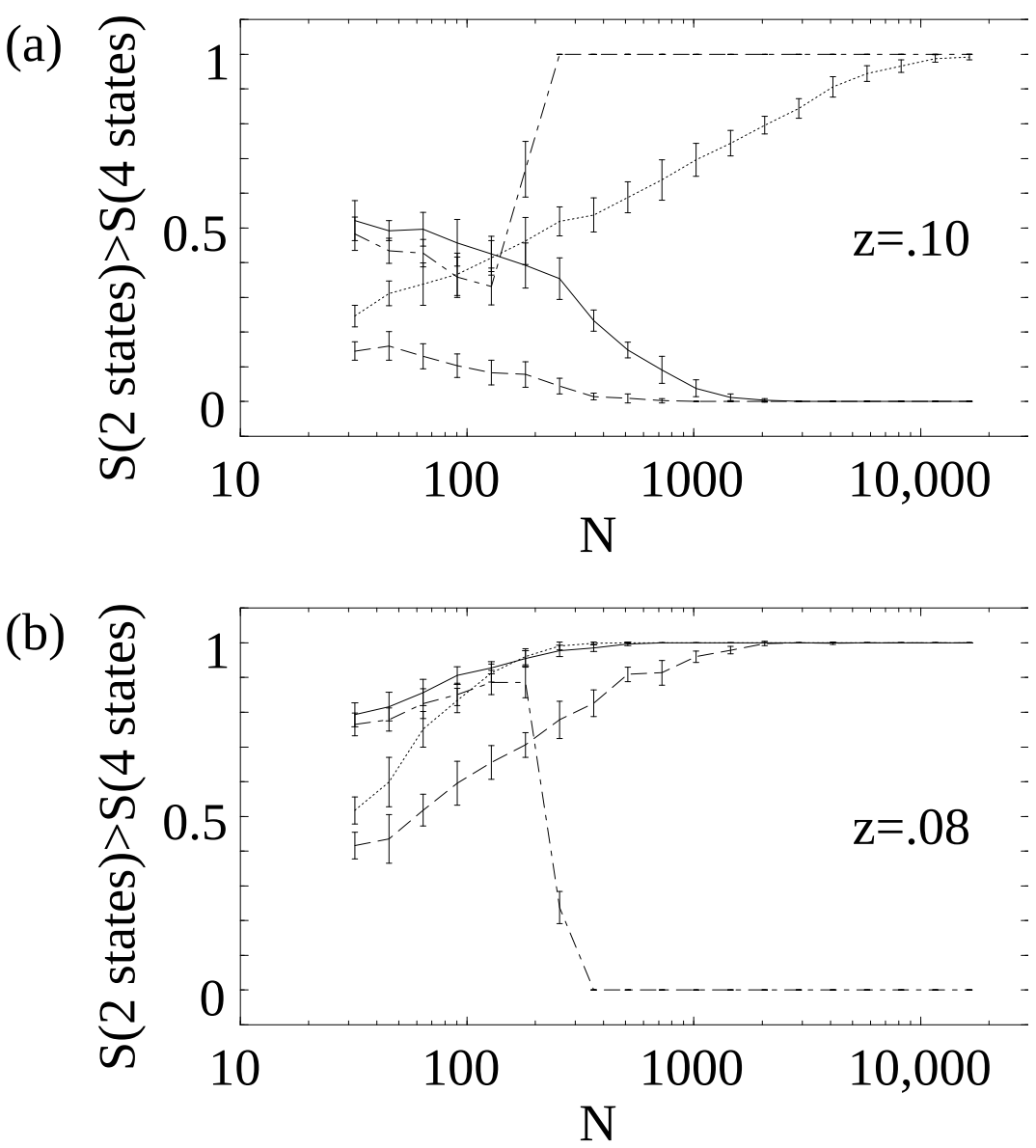

Figure 3: Variable selection: the y-axis shows the fraction where the dependence-measure favours a variable with 2 states over one with 4 states, based on 100 resamples for each $N$; $N = 32, ..., 16384$ along the x-axis. The four measures are $\hat{I}^{\rm BC}$ (dashed line), $\widehat{SI}$ (solid), $\widehat{NI}$ (dotted) and p-value (dash-dotted).

Concerning variable/feature selection (cf. [8, 13] for an overview), our second experiment illustrates the benefits of $\widehat{SI}$. Our simplistic aim is to find a *single* variable that predicts the class label best, rather than determining the optimal set of variables.

We chose a simple gold-standard model: the class variable $Y$ is the root of a naive Bayes net with 20 leaf nodes $X_1, ..., X_{20}$. $Y$ takes each of its four states with equal probability. The ten variables $X_1, ..., X_{10}$ are binary, each of which being in state 1 with probabilities .6, .8, .3 and .1 conditioned on $Y$ being in states 1, 2, 3 and 4, respectively.[5] The other ten variables $X_{11}, ..., X_{20}$ have four states; each of these variables is in the same state as $Y$ with probability $(1/4 + 3z)/4$ and in any other state with probability $(1/4 - z)/4$; parameter $z \in [0, 1/4]$ determines the degree of dependence between $Y$ and the variables with four states.

From this model, we sampled data sets of various sizes and for different values of $z$. When $z \approx .0882$, we have for the *true* mutual information: $I(Y, X_i) = I(Y, X_j)$ for all $i, j = 1, ..., 20$. Fig. 3a depicts the case where $z = .10 > .0882$, i.e., the variables with four states provide more *true* mutual information about $Y$ than the binary ones do. Fig. 3b shows the results for the opposite case, where $z = .08 < .0882$.

In Figure 3a, our new measure $\widehat{SI}$ and the bias-corrected mutual information clearly favour a variable with four states when $N$ is large, as expected. The bias-corrected information does so even for small sample sizes (erroneously). In contrast, our standardized information $\widehat{SI}$ as well as the p-value favour the simpler hypothesis, namely 2 states, at small sample sizes, as desirable (e.g., Occam's razor). Note that the computation of the p-value again suffered from limited numerical accuracy at large $N$, as before.[6]

At small sample sizes $N$, the normalized mutual information $\widehat{NI}$ gives reasonable results again, as its values are between the ones of the p-value and $\hat{I}^{\rm BC}$ (in Figures 3a and b). The fact that the regularization of $\widehat{NI}$ does not decrease with growing $N$ (cf. Section 5) favours the variables with 2 states over the ones with four states at large $N$. This explains the (erroneous) increase of the curve pertaining to $\widehat{NI}$ in Fig. 3a, as well as the steep increase in Fig. 3b.

In Fig. 3b, all the measures (except for the numerically instable p-value) agree on a variable with 2 states for large $N$, as expected. Even though the sampling noise at small $N$ can induce a higher score for a variable with 4 states by chance, our new standardized information $\widehat{NI}$ as well as the p-value appear quite robust. In contrast, the bias-corrected information seems to be quite sensitive to this noise, favouring variables with a larger number of states at small $N$, which contradicts the idea of regularization (e.g., Occam's razor).

### 7.2 Equivalent Sample Size

We determined the optimal value of the equivalent sample size $N'$ based on our Eqs. 9 and 10 for the data by Sewell and Sha [17] and the alarm-network data [2]. Figure 4 shows that our linear approximation of Eq. 9 indeed is valid for the data by Sewell et al. [17]. It also yields that $N' \approx 67$. This is close to $N' \approx 66.1$ obtained from our analytical approximation (Eq. 10). These results agree with the more involved alternatives: empirical Bayes results in $N' \approx 69$, and cross validation yields $N' \approx 100, ..., 300$ [19].

[5]We found that the qualitative behavior of our results does not depend on the particular choice of these probabilities, as expected.

[6]When it happens, we on purpose choose the 'wrong' result as to display clearly the ample size $N$ at which the numerical instability of the p-value occurs in the figure.

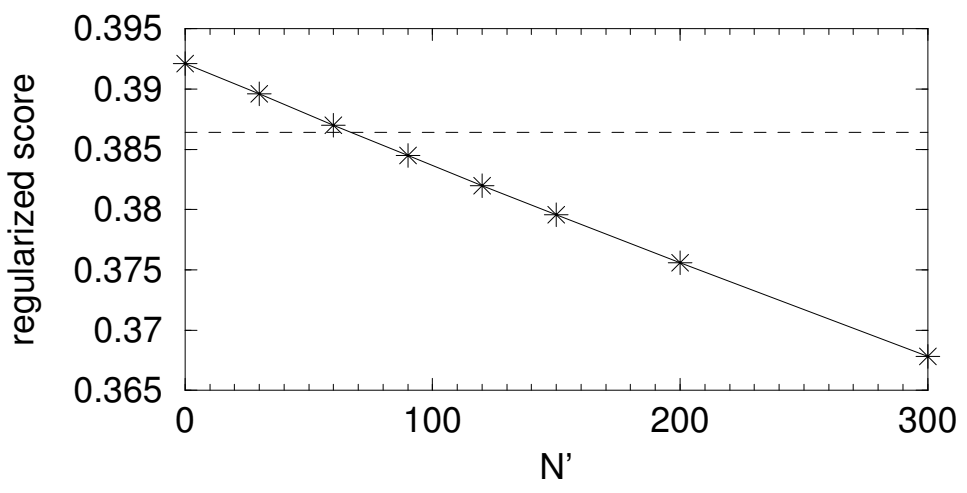

Figure 4: The value of the right (dotted line) and left (solid) hand side of Eq. 9 for various values of $N'$; data by Sewell et al. [17], $N = 10318$.

Given data of size $N = 10000$ sampled from the alarm network [2], our constraint in Eq. 9 yields $N' \approx 12$, while our analytical approximation in Eq. 10 results in $N' \approx 14.7$. The increased value of our approximation is due to zero cell counts in the data. These results agree with $N' \approx 16$ found in [9] using cross-validation and structural difference.

Compared to the data by Sewell et al. [17], $N'$ is smaller for the alarm network, as expected: the parameters in the alarm-network take quite extreme values, resulting in a large value of $\hat{I}$ (strong dependence), which appears in the denominator of Eq. 10.

## 8 Conclusions

We have proposed two conditions for regularization of dependence-measures, which apply to the extreme cases of vanishing and strong dependence. This led us to a simple yet effective new estimator with interesting properties. While established measures failed already in simplistic experiments, our measure appeared quite robust. Moreover, our approach also shed light on the applicability of the normalized mutual-information, which is commonly used in applications but theoretically not well understood. Along these lines, we also obtained a simple constraint for regularizing parameter estimates in graphical models. We showed that, for the equivalent sample size, our linear approximation is valid and computationally efficient.

### Acknowledgements

I am grateful to R. Bharat Rao for encouragement and support of this work, and to the anonymous reviewers for useful comments.